\documentclass{article}

\usepackage[preprint]{neurips_2026}
\usepackage[utf8]{inputenc}
\usepackage[T1]{fontenc}
\usepackage{hyperref}
\usepackage{url}
\usepackage{booktabs}
\usepackage{amsfonts}
\usepackage{amsmath}
\usepackage{nicefrac}
\usepackage{microtype}
\usepackage{graphicx}
\usepackage{xcolor}
\usepackage{caption}
\usepackage{enumitem}
\graphicspath{{figures/}}
\hypersetup{
  pdftitle={The Geometry of Agent Memory: Forecasting Retrieval-Bound Memory Failures, and Characterizing Its Limits},
  pdfauthor={Iliya Zhechev; Radoslav Ralev; Chris Coleman; Srijith Rajamohan; Jen Agarwal; Tyler Hutcherson; Simba Khadder},
  pdfsubject={PALM Workshop at NeurIPS 2026},
  pdfkeywords={agent memory, dense retrieval, embedding geometry, correctness prediction}
}

\title{The Geometry of Agent Memory:\\ Forecasting Retrieval-Bound Memory
Failures, and Characterizing Its Limits}

\author{%
  Iliya Zhechev\textsuperscript{1} \quad Radoslav Ralev\textsuperscript{2} \quad Chris Coleman\textsuperscript{3}\\
  \textbf{Srijith Rajamohan\textsuperscript{4} \quad Jen Agarwal\textsuperscript{5} \quad Tyler Hutcherson\textsuperscript{6}}\\
  \textbf{Simba Khadder\textsuperscript{7}}\\
  Redis\\
  {\small\textsuperscript{1}\texttt{iliya.zhechev@gmail.com}\quad
    \textsuperscript{2}\texttt{radoslav.ralev@redis.com}}\\
  {\small\textsuperscript{3}\texttt{ccol@u.northwestern.edu}\quad
    \textsuperscript{4}\texttt{srijith.rajamohan@gmail.com}}\\
  {\small\textsuperscript{5}\texttt{honey.agarwal@gmail.com}\quad
    \textsuperscript{6}\texttt{tyler.hutcherson@redis.com}}\\
  {\small\textsuperscript{7}\texttt{simba.khadder@redis.com}}
}

\begin{document}
\maketitle

\begin{center}
  \small Accepted as a poster at the
  \href{https://openreview.net/forum?id=xP0xvmQZhB}{PALM Workshop at NeurIPS 2026}.
\end{center}

\begin{abstract}
We study whether geometric relationships between conversation turns, memory
texts, and queries can predict answer correctness before generation.
We evaluate four memory implementations with dense or hybrid retrieval and one
observation-summary comparator. We embed these texts in a shared space.
A logistic classifier
uses 25 features describing query proximity, memory--conversation
correspondence, and conversation structure.
On LongMemEval-small, pooled AUROC is 0.640 over 2{,}367 question--system pairs,
below a question-difficulty baseline of 0.655. Performance varies by question
type: geometry reaches 0.841 on single-session-assistant questions, but system
identity alone reaches 0.826. After controlling for difficulty and identity,
geometry adds no statistically established pooled improvement
($\Delta=+0.002$, 95\% CI $[-0.010,0.014]$).
A fixed-store context-selection experiment finds higher accuracy with
proximity-ranked memories than with random subsets, consistent with ordinary
retrieval relevance. On LoCoMo, nearest-turn similarity has modest predictive
value for single-hop questions and near-chance estimates for several other
categories. Training uses correctness labels; prediction uses pre-answer
features and no additional LLM generation call.
The results characterize associations between memory representations and
correctness, rather than establish a reliable deployment diagnostic.
In particular, embedding proximity does not verify fact coverage or distinguish
memory-construction, retrieval, and generation errors.
\end{abstract}

\section{Introduction}

Long-term agent memory can use text retrieval, summaries, knowledge graphs,
symbolic stores, or parametric updates. This study focuses on text memory
representations associated with dense retrieval. We also include an
observation-summary system as a comparator. Benchmark QA accuracy measures one
aspect of these systems, alongside cost, latency, storage, and other concerns.
A separate question is whether a signal available before generation can help
identify queries that are likely to receive an incorrect answer.

\begin{figure}[t]
  \centering
  \includegraphics[width=\linewidth]{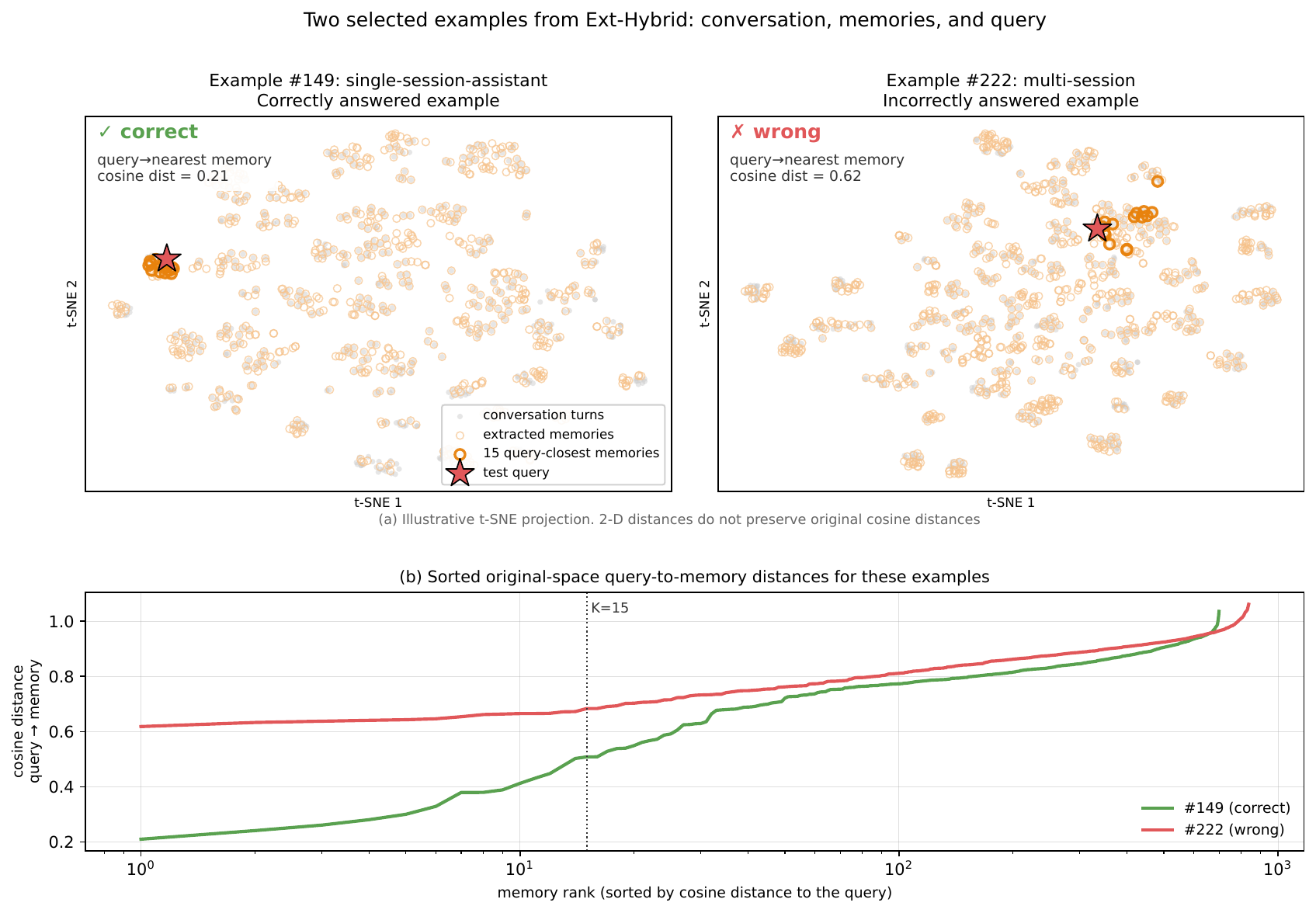}
  \caption{Conversation turns, memories, and queries for two selected examples.
  The correctly answered query is closer to memory points than the incorrectly
  answered query. The t-SNE projection is illustrative; all measurements use
  original-space cosine distance. The examples do not establish fact coverage
  or the cause of an error.}
  \label{fig:manifold}
\end{figure}

We study three representations: source conversation turns, memory texts produced
by a system, and an incoming query. Dense retrieval systems already embed
memories and queries. We additionally embed conversation turns in the same space
(Figure~\ref{fig:manifold}). We compute two feature blocks and conversation covariates:
\begin{itemize}[leftmargin=1.3em,itemsep=1pt,topsep=2pt]
\item \textbf{Query reachability}: nearest-memory distance, mean distance to the
  five closest memories, local memory density, and nearest-turn distance.
  Here, reachability names an embedding-proximity measurement.
\item \textbf{Memory--conversation correspondence}: turn-to-memory proximity,
  session alignment, memory count, and memory dispersion. These describe
  geometric correspondence rather than verified coverage of individual facts.
\end{itemize}

A logistic classifier predicts answer correctness from these features.
We examine variation across question types, incremental prediction beyond
question difficulty and system identity, and accuracy under different
context-selection rules.
The strongest observed AUROC is 0.841 for single-session-assistant questions.
However, system identity alone reaches 0.826 on that same type.
The high within-type score therefore does not establish independent prediction
of instance-level memory failures.
Pooled over all types, geometry adds no significant gain beyond difficulty and
system identity (\S\ref{sec:general}).

Memory construction can omit evidence, retrieval can miss evidence already
stored, and generation can fail despite sufficient retrieved context.
We test the hypothesis that the first two cases leave geometric signatures.
Our use of \emph{retrieval-bound} refers to this motivating hypothesis, not a
verified classification of the observed errors. Distances alone cannot show
whether an answer-bearing fact was stored, included in the actual prompt, or
used correctly.

A warning score could eventually support abstention, broader retrieval, or
recovery from the source conversation. Our present results do not validate such
a deployment policy. The scope analysis uses labeled outcomes, and prediction
across unseen memory architectures remains unreliable.

\noindent Analysis code and numerical artifacts:
\href{https://github.com/ichko/agent-memory-geometry}{github.com/ichko/agent-memory-geometry}.

\section{Related work}

\paragraph{Agent-memory benchmarks and systems.}
LongMemEval~\citep{wu2025longmemeval} evaluates 500 questions in six categories
over multi-session histories. LoCoMo~\citep{maharana2024locomo} uses a different
conversational dataset and question taxonomy.
Memory-system studies report end-to-end QA results alongside other
measurements~\citep{chhikara2025mem0,rasmussen2025zep}.
We examine whether features of the memory representation are associated with
per-question correctness.

\paragraph{Selective prediction and confidence.}
Selective prediction ranks examples by expected correctness and allows
deferral~\citep{hendrycks2017baseline,geifman2017selective}.
Verbalized confidence and P(True) use the answer model's own
judgment~\citep{kadavath2022know}. Our geometric score is supervised during
training but does not require another generation call at prediction time.

\paragraph{Retrieval diagnostics and query-performance prediction.}
Query--passage similarity is an established retrieval signal.
Query-performance prediction estimates retrieval difficulty from query or
ranked-list statistics~\citep{cronentownsend2002predicting,carmel2010estimating}.
RAG evaluation also measures answer groundedness, often using LLM
judges~\citep{es2024ragas,saadfalcon2024ares}.
Our reachability features overlap with ordinary similarity and ranked-list
signals. The setting here is a memory representation constructed from a
conversation. We add conversation--memory correspondence, compare correctness
across memory implementations, and examine differences by question type.
The intervention tests the value of relevant context, without establishing
a new geometric cause separate from relevance.

\section{Geometric features}
\label{sec:features}

A shared model $E:\text{text}\to\mathbb{R}^d$ embeds conversation turns $T$,
stored memory texts $M$, and a query $q$. Distances are cosine distances between
unit-normalized embeddings. The core reachability scalar is
$\texttt{q\_nn\_mem}=\min_{m\in M}d(q,m)$.
The query block has six features. The correspondence block has 14 features:
seven coverage, two alignment, and five memory-dispersion features.
The full model also includes five conversation-structure covariates,
for 25 features in total (Appendix~\ref{app:features}).

Coverage here means proximity between embedded texts.
A nearby memory may omit the required detail, express a stale fact, or contradict
the source. A distant memory may still contain the answer.
The classifier is logistic regression on standardized scalar features.
Inspectable inputs do not give its coefficients a causal interpretation.

\section{Experimental setup}
\label{sec:setup}

\paragraph{Systems and benchmark.}
We evaluate five implementations on LongMemEval-small.
Four retrieve text using dense similarity, with hybrid retrieval in one case.
The fifth constructs an observation summary and passes it to the answer model.
We use the following functional names:
\begin{itemize}[leftmargin=1.2em,itemsep=1pt,topsep=2pt]
\item \textbf{Sem-RAG}: semantic-chunk RAG over source turns (65.8\% accuracy).
\item \textbf{Obs-Extract}: a local Mastra-style observation-memory implementation
  (67.9\%). Its logged representation is one observation summary.
\item \textbf{Ext-Dense}: instruction-based extraction and dense retrieval (72.6\%).
\item \textbf{Svc-Mem}: a local LangMem-based implementation (73.2\%).
\item \textbf{Ext-Hybrid}: instruction-based extraction and hybrid retrieval (86.0\%).
\end{itemize}
These results concern the particular configurations in our harness.
They do not represent all configurations of the named libraries or products.
The answer model and judge use the \texttt{gpt-4o} API alias across these runs.
Answer prompts remain specific to each implementation.
The feature analysis uses deduplicated memory texts recorded in answer logs,
rather than the precise context passed to the answer model.
Joining these representations and judged answers gives 2{,}367
question--system pairs over 500 distinct questions.
Appendix~\ref{app:repro} gives counts and implementation details.

\paragraph{Protocol.}
We use \texttt{StratifiedGroupKFold} with five folds, grouping by
\texttt{question\_id} and using seed 0. This keeps all system rows for a question
in one fold. Scaling is fitted on each training fold.
Reported AUROCs use one cross-validation pass.
Each per-type result comes from a separate classifier trained within that type.
We predict correctness and report AUROC for that ranking; reversing labels and
scores gives the corresponding failure AUROC.
95\% intervals resample question groups from fixed out-of-fold predictions,
with 1{,}000 draws. They capture sampling variation conditional on that fit,
not model-refitting or split-selection uncertainty.
Paired comparisons use the same question-group resampling.

The original analysis imputes undefined features using column means before
the fold split. The question-difficulty baseline likewise fits unsupervised
PCA on the full question set. These operations do not use labels, but expose
preprocessing to held-out covariates.
Grouping controls the specific leakage caused by repeated question rows.
The conversation-only baseline is 0.505 under this grouped protocol.
Primary embeddings use \texttt{text-embedding-3-small}.

\section{Results}
\label{sec:results}

\subsection{Prediction varies by question type}
\label{sec:pertype}

\begin{figure}[t]
  \centering
  \includegraphics[width=0.82\linewidth]{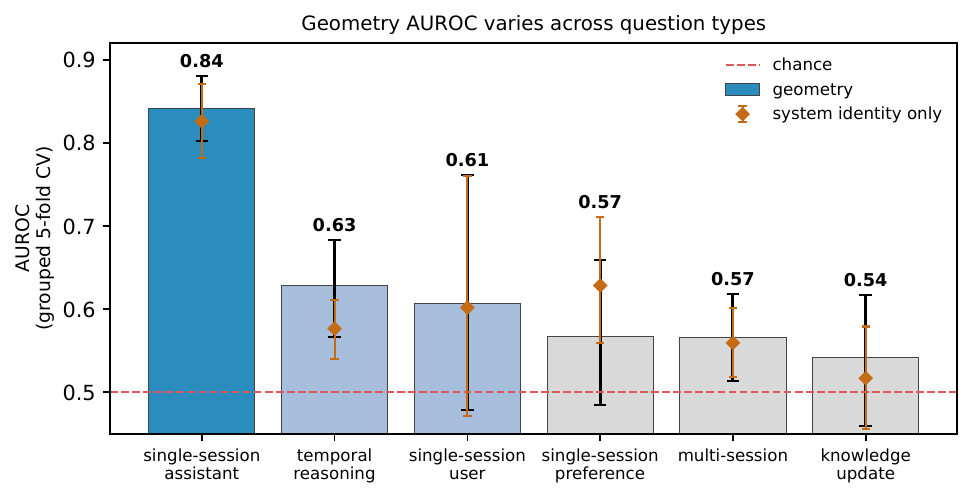}
  \caption{AUROC within each type (five implementations). Bars show full geometry;
  diamonds show a system-identity-only classifier under the same protocol.
  Intervals are question-group bootstrap 95\% intervals conditional on fixed
  out-of-fold predictions.}
  \label{fig:pertype}
\end{figure}

Pooled geometry AUROC is 0.640 $[0.615,0.664]$ over 2{,}367 rows.
Table~\ref{tab:pertype} and Figure~\ref{fig:pertype} show variation across types.
Single-session-assistant has the strongest score, 0.841,
whereas temporal reasoning reaches 0.628.
Preference and knowledge-update are close to chance.
The multi-session score is modestly above chance, with an interval excluding 0.5.

\begin{table}[h]
  \centering\small
  \caption{Geometry AUROC by type, with separately fitted classifiers and
  question-group bootstrap intervals. $n$ counts question--system pairs.
  These results do not classify the causes of individual errors.}
  \label{tab:pertype}
  \begin{tabular}{lrrlr}
    \toprule
    Question type & $n$ & AUROC & 95\% CI & \% wrong \\
    \midrule
    single-session-assistant & 265 & \textbf{0.841} & $[0.802, 0.881]$ & 27.5 \\
    temporal-reasoning        & 623 & 0.628 & $[0.567, 0.683]$ & 27.8 \\
    single-session-user       & 333 & 0.607 & $[0.478, 0.762]$ & \phantom{0}8.1 \\
    single-session-preference & 137 & 0.567 & $[0.485, 0.659]$ & 48.2 \\
    multi-session             & 639 & 0.566 & $[0.514, 0.618]$ & 33.3 \\
    knowledge-update          & 370 & 0.542 & $[0.459, 0.617]$ & 21.1 \\
    \bottomrule
  \end{tabular}
\end{table}

Within-type evaluation holds question type constant, but it does not hold
system identity or question content constant. For single-session-assistant,
identity alone reaches 0.826. Geometry's paired improvement is
$+0.015$ $[-0.020,0.049]$ and is not statistically established.
Memory count alone reaches 0.830. The strongest score is therefore compatible
with representation-size and system-accuracy differences
(\S\ref{sec:general}). The single-session-user estimate is particularly
uncertain because only 27 of its 333 rows are failures.

\subsection{Feature-block ablation}
\label{sec:decomp}

Table~\ref{tab:decomp} compares the correspondence and query blocks with the
full 25-feature model. Their pooled AUROCs are 0.580, 0.598, and 0.640.
The full model also includes conversation-structure covariates,
so the comparison does not isolate a correspondence--query interaction.

Full geometry exceeds the query block by $+0.042$ $[0.019,0.065]$
and correspondence by $+0.060$ $[0.036,0.085]$ in paired
question-group bootstrap comparisons.
The query-minus-correspondence difference is $+0.018$ $[-0.013,0.049]$,
so their numerical ordering is not statistically established.
Within types, neither block consistently dominates.
For single-session-assistant, correspondence alone reaches 0.843.
For preference and single-session-user, the full model underperforms
correspondence alone. The pooled ordering does not describe every type.

\begin{table}[h]
  \centering\small
  \caption{Feature-block AUROCs under grouped CV. Full geometry includes both
  blocks and five conversation covariates. Paired pooled comparisons are
  reported in the text; per-type block comparisons are point estimates.
  $^{*}$Single-session-user has few errors and substantial uncertainty.}
  \label{tab:decomp}
  \begin{tabular}{lrrr}
    \toprule
    Question type & memory--conv.\ correspondence & query reachability & full geometry \\
    \midrule
    single-session-assistant  & 0.843 & 0.825 & 0.841 \\
    single-session-preference & \textbf{0.605} & 0.478 & 0.567 \\
    temporal-reasoning        & 0.590 & 0.612 & 0.628 \\
    multi-session             & 0.562 & 0.517 & 0.566 \\
    knowledge-update          & 0.547 & 0.512 & 0.542 \\
    single-session-user$^{*}$ & 0.645 & 0.475 & 0.607 \\
    \midrule
    pooled                    & 0.580 & 0.598 & 0.640 \\
    \bottomrule
  \end{tabular}
\end{table}

\subsection{The reachability gap is an offline description}
\label{sec:scope}

For each type, the \emph{reachability gap} is the mean nearest-memory distance
for incorrect answers minus that for correct answers.
Single-session-assistant has a substantially larger positive gap than other
types (Table~\ref{tab:gap}).
This describes an association between proximity and correctness.
It does not identify whether the required fact was absent from the store,
absent from the prompt, or incorrectly used.
Similar mean distances for knowledge-update and preference do not imply
identical distance distributions or rule out other predictive features.

\begin{table}[h]
  \centering\small
  \caption{Mean nearest-memory cosine distances by judged correctness.
  AUROC is from the full per-type classifier in Table~\ref{tab:pertype}.
  The large-gap category has the highest AUROC, but the smaller-gap categories
  do not show a precise monotonic ordering.}
  \label{tab:gap}
  \begin{tabular}{lrrrr}
    \toprule
    Question type & correct & wrong & \textbf{gap} & AUROC \\
    \midrule
    single-session-assistant  & 0.377 & 0.618 & \textbf{$+$0.241} & 0.841 \\
    single-session-user       & 0.533 & 0.579 & $+$0.046 & 0.607 \\
    temporal-reasoning        & 0.526 & 0.570 & $+$0.044 & 0.628 \\
    multi-session             & 0.534 & 0.567 & $+$0.033 & 0.566 \\
    knowledge-update          & 0.456 & 0.454 & $-$0.003 & 0.542 \\
    single-session-preference & 0.570 & 0.541 & $-$0.029 & 0.567 \\
    \bottomrule
  \end{tabular}
\end{table}

Gaps are computed before rounding the mean distances.

For knowledge-update, the full classifier reaches 0.542.
Exploratory nonlinear and raw-embedding probes did not establish a reliable
memory-specific improvement on this category.
Selecting a stale fact is one possible explanation. We have not audited
the actual evidence sufficiently to identify that cause.

The gap uses correctness labels and describes the evaluated data offline.
It cannot tell a deployed agent whether an individual prediction is trustworthy.
It concerns nearest-memory distance rather than the full feature set.
Preference has a negative gap while correspondence-only reaches 0.605,
illustrating that distinction.

\subsection{Reference-answer proximity: a proxy analysis}
\label{sec:decompose}

We embed the reference answer as an offline anchor.
Let $m^*$ be its nearest memory, $c=d(a,m^*)$ its distance to reference answer $a$,
and $r=d(q,m^*)$ its distance to the query.
Thresholds $\tau_c$ and $\tau_r$ are the 75th percentiles of these quantities
among correctly answered rows.
An incorrect row is \emph{answer-distant} if $c>\tau_c$,
\emph{query-distant} if $c\leq\tau_c$ and $r>\tau_r$, and
\emph{neither} otherwise.

These bins do not identify extraction, retrieval, or reasoning errors.
Proximity cannot establish answer entailment, and $r$ does not show
whether $m^*$ reached the actual prompt.
Table~\ref{tab:decompose} retains the measurements with operational names.

\begin{table}[h]
  \centering\small
  \caption{Reference-answer proximity among 630 failed rows.
  Thresholds are the 75th percentiles among correct rows.
  These are proxy bins, rather than verified extraction, retrieval, or
  reasoning failure rates. Percentages are rounded independently.}
  \label{tab:decompose}
  \begin{tabular}{lrrrr}
    \toprule
    Question type & \# fail & answer-distant & query-distant & neither \\
    \midrule
    single-session-assistant  & \phantom{0}73 & \textbf{62\%} & \phantom{0}5\% & 33\% \\
    multi-session             & 213 & 37\% & 21\% & 43\% \\
    knowledge-update          & \phantom{0}78 & 31\% & \phantom{0}8\% & \textbf{62\%} \\
    temporal-reasoning        & 173 & 25\% & 13\% & \textbf{61\%} \\
    single-session-user       & \phantom{0}27 & 19\% & 19\% & 63\% \\
    single-session-preference & \phantom{0}66 & \phantom{0}0\% & \phantom{0}3\% & \textbf{97\%} \\
    \midrule
    all                       & 630 & 31\% & 13\% & 56\% \\
    \bottomrule
  \end{tabular}
\end{table}

For single-session-assistant, 62\% of failed rows are answer-distant and
5\% are query-distant. The 62\% does not mean the fact was verified as absent.
None of the 2{,}367 evaluated memory representations is empty.
In Obs-Extract, however, the whole summary is represented as one text,
changing the granularity of the comparison.

At the 90th-percentile thresholds, the pooled \emph{neither} share
increases from 56\% to 77\%.
The reference anchor and correctness-conditioned thresholds couple this
analysis to the labels.
We have not measured separate AUROCs for verified
\emph{fact absent} and \emph{fact present but not retrieved} cases.
That requires checking answer entailment in the store and actual prompt,
including current and stale facts for knowledge-update questions.

\subsection{Proximity-ranked context selection affects accuracy}
\label{sec:causal}

On Ext-Hybrid, we keep the memory pool fixed and vary the subset provided
to the answer model: 15 cosine-closest memories, the native ranking,
random subsets over five seeds, or 15 farthest memories.
Table~\ref{tab:causal} reports accuracy under two answer-model aliases.

\begin{table}[h]
  \centering\small
  \caption{Interventional accuracy when the answer model is given different memory
  subsets from the \emph{same} store (Ext-Hybrid, $K{=}15$, two answer models).}
  \label{tab:causal}
  \begin{tabular}{lrr}
    \toprule
    Memories shown to the model & gpt-4o-mini & gpt-4o \\
    \midrule
    Farthest-15 (cosine floor)         & 0.068 & 0.060 \\
    Random-15 (5 seeds)                & 0.114 & 0.091 \\
    Closest-15 (pure cosine)           & 0.752 & 0.790 \\
    Native-15 (system's own retriever) & 0.738 & 0.822 \\
    \bottomrule
  \end{tabular}
\end{table}

Closest-15 reaches 0.752 with \texttt{gpt-4o-mini}, compared with
0.114 for random selection. With \texttt{gpt-4o}, the values are 0.790 and 0.091.
Native ranking reaches 0.738 and 0.822.
The ordering between native and cosine selection depends on the model.

The intervention shows that selected context affects answer accuracy.
Cosine ranking is an ordinary relevance criterion.
The experiment does not isolate reachability from relevance, establish causal
effects of all 25 features, or show that the classifier improves retrieval.
Fixed memory counts can also produce different context token lengths.

\subsection{Baselines and system confounding}
\label{sec:general}

\paragraph{Pooled baselines.}
Geometry (0.640) exceeds nearest-memory similarity alone (0.603),
with paired $\Delta=+0.037$ $[0.012,0.063]$.
Question difficulty uses type, length, and PCA-64 of the question embedding.
Its AUROC is 0.655. Adding geometry gives 0.681, an improvement of
$+0.026$ $[0.010,0.042]$ over difficulty alone.

The pooled increment is not established after adding system identity.
Identity alone reaches 0.578.
Difficulty plus identity reaches 0.679, while adding geometry gives 0.681.
The paired difference is $+0.002$ $[-0.010,0.014]$.
The evaluated geometry therefore does not establish additional pooled
prediction beyond these controls.

\paragraph{The strongest type also has a system confound.}
Within single-session-assistant, identity alone reaches 0.826
$[0.782,0.871]$, compared with geometry at 0.841 $[0.802,0.881]$.
The paired difference is $+0.015$ $[-0.020,0.049]$.
Memory count alone reaches 0.830.
Obs-Extract accounts for 41 of this type's 73 errors,
whereas Ext-Hybrid answers all 56 evaluated questions of this type correctly.
Type stratification alone does not remove the architecture confound.

\paragraph{Within-system prediction and held-out systems.}
Separately trained within-system geometry models have modest AUROCs:
Sem-RAG 0.657, Ext-Dense 0.597, Ext-Hybrid 0.635,
Obs-Extract 0.552, and Svc-Mem 0.580.
Training on four systems and testing on the fifth gives a mean AUROC of 0.603,
with 0.467 on one held-out system.
Those fits reuse question identities across training and test systems.
They test an unseen system on overlapping questions, rather than joint
transfer to new systems and questions.

\paragraph{Other signals and embedding models.}
For three systems with confidence elicitation, verbalized confidence
has AUROC about 0.43 and P(True) 0.43 $[0.393,0.466]$ ($n=1{,}368$).
This indicates reversed ranking in this setting.
AUROC alone does not establish probability miscalibration.

Recomputing geometry gives pooled AUROCs
0.640 (\texttt{text-embedding-3-small}),
0.645 (\texttt{text-embedding-3-large}),
0.626 (\texttt{BAAI/bge-small-en-v1.5}), and
0.621 (\texttt{thenlper/gte-base}).
Their mean is 0.633, with sample standard deviation 0.012 across models.
This checks pooled performance. It does not establish stability of every
per-type or per-system result across embeddings.

\subsection{Selective answering: an exploratory illustration}
\label{sec:selective}

Ranking original out-of-fold predictions by P(correct) and retaining the
top-scoring fraction raises accuracy among retained rows from 0.734 at
full coverage to 0.786 at 70\% coverage and 0.841 at 30\%.
These are retrospective calculations on the evaluation predictions,
rather than a separately tuned and tested deferral policy.
They do not establish a benefit over difficulty-plus-identity deferral,
and no intervals for these coverage points are reported.
The score requires no additional LLM generation call, but embeddings and
feature computation still incur cost.

\subsection{Nearest-turn similarity on LoCoMo}
\label{sec:locomo}

The LoCoMo analysis uses conversation turns as candidate memories,
without a separate extraction step.
With \texttt{gpt-4o-mini} generation and judging, Closest-15 accuracy
is 0.501, compared with Random-15 at 0.051 and Farthest-15 at 0.003
($n=1{,}986$ per condition).
This reproduces the proximity-ranked versus random-context ordering.

Table~\ref{tab:locomo} reports AUROC for the fixed score
$-\min_{t\in T}d(q,t)$ against full-context answer labels.
No logistic classifier is fitted in this category analysis, and no trained
LongMemEval predictor is transferred.
Single-hop reaches 0.603, while multi-hop, temporal, and open-domain have
near-chance point estimates. Adversarial questions reach 0.612.
Answerability can affect query relevance in that category, so we treat it as
descriptive rather than separate mechanistic evidence.

The single-hop score is consistent with a weaker version of the LongMemEval
association. We have only 10 conversations, and do not report
conversation-level confidence intervals for the category scores.
The comparison does not establish a common cause for the errors.

\begin{table}[h]
  \centering\small
  \caption{LoCoMo category results for a fixed nearest-turn similarity score.
  $n$ counts questions. Gaps use full-context answer correctness.
  These descriptive point estimates have no category confidence intervals.}
  \label{tab:locomo}
  \begin{tabular}{lrrrr}
    \toprule
    LoCoMo category & $n$ & \% wrong & gap & AUROC \\
    \midrule
    single-hop            & 841 & 17.7 & $+$0.026 & 0.603 \\
    adversarial (no answer)& 446 & 62.6 & $+$0.032 & 0.612 \\
    multi-hop             & 282 & 43.3 & $-$0.001 & 0.520 \\
    open-domain           & 96 & 65.6 & $-$0.003 & 0.487 \\
    temporal              & 321 & 81.3 & $-$0.007 & 0.468 \\
    \bottomrule
  \end{tabular}
\end{table}

\section{Discussion}
\label{sec:discussion}

\paragraph{Interpretation.}
Proximity and representation features are associated with correctness,
and the association varies by question type.
However, system identity alone approaches the high within-type score,
and geometry does not add significant pooled value beyond identity and difficulty.
The intervention confirms the importance of selected context,
consistent with established retrieval relevance.

For memory-system evaluation, this distinction matters: a high pooled or
within-type prediction score can reflect differences between implementations,
rather than detect unsupported queries within an implementation.
Comparisons should therefore include system-identity and question-difficulty
controls, alongside results within comparable memory representations.

The reachability gap summarizes outcomes offline.
It does not certify evidence sufficiency or a live prediction's reliability.
An answer-distant reference anchor suggests a possible memory omission,
but cannot verify one. A useful next study would audit whether an
answer-entailing memory exists in the store and reaches the actual prompt.
That could separate memory construction, retrieval, and generation errors
more directly than embedding distances.

\paragraph{Limitations.}
The study covers two English benchmarks and five harness implementations.
Obs-Extract exposes a whole summary as one memory text, so representation
granularity differs substantially across systems.
Question type does not control architecture or question-content effects.
The analysis uses one grouped-CV split and fixed-prediction bootstrap intervals.
Full-set imputation and question PCA expose preprocessing to held-out covariates.
Feature choices were explored on this benchmark without a final independent
test set or nested model selection.
The intervention uses one extracted-memory system and fixed memory counts
rather than equal token budgets.
The reference-anchor bins use labels and thresholds, with no entailment audit.
LoCoMo category estimates lack conversation-level uncertainty.
API aliases were not recorded as dated model snapshots.
These limits preclude claims of general deployment reliability.

\paragraph{Privacy and broader impacts.}
A validated diagnostic might help reduce unsupported answers through deferral
or additional evidence gathering. An unreliable score could instead encourage
false confidence or unnecessarily withhold useful answers.
We have not evaluated fairness across user groups or languages.
Correspondence features require access to source-turn embeddings.
Deployments should enforce consent, retention limits, joint deletion of turns
and derived embeddings, and access controls between users.
Memory poisoning and disclosure to external embedding providers also require
consideration. We evaluate existing benchmarks, without introducing
a study with human participants.

\paragraph{Conclusion.}
Geometry describes memory representations and their association with answer
correctness. The observed association is strongest for single-session-assistant,
but system differences account for much of that result.
The incremental pooled signal beyond difficulty and identity is not established.
The contribution is a characterization of the evaluated signals and their limits.
Verified failure attribution and deployment decisions require further study.

\bibliographystyle{plainnat}
\bibliography{references}

\newpage
\appendix
\section{The 25 geometric features}
\label{app:features}

All features use the embedding space over the turn set $T$ (with session labels),
the memory set $M$, and the query $q$; distances are cosine. $\varepsilon_p$ is the
$p$-th percentile of pairwise turn distances (a per-example, adaptive scale).

\begin{table}[h]
  \centering\small
  \caption{The 25 scalar features, grouped by the two relationships of \S1.}
  \label{tab:features}
  \begin{tabular}{p{0.30\linewidth}p{0.63\linewidth}}
    \toprule
    Family / feature & Definition \\
    \midrule
    \multicolumn{2}{l}{\emph{Correspondence: coverage (7)}}\\
    \texttt{cov\_mean\_nn}, \texttt{cov\_max\_nn} & mean / worst nearest-memory distance over turns \\
    \texttt{cov\_hausdorff} & Hausdorff$(T,M)$ \\
    \texttt{cov\_mean\_session}, \texttt{cov\_worst\_session} & mean / worst per-session coverage \\
    \texttt{cov\_at\_eps\_p10}, \texttt{cov\_at\_eps\_p5} & fraction of turns with a memory within $\varepsilon_p$ \\
    \midrule
    \multicolumn{2}{l}{\emph{Correspondence: alignment (2)}}\\
    \texttt{mem\_silhouette} & silhouette of $M$ under nearest-turn session labels \\
    \texttt{bridging\_fraction} & fraction of memories whose two nearest turns are in different sessions \\
    \midrule
    \multicolumn{2}{l}{\emph{Memory dispersion (5)}}\\
    \texttt{mem\_count} & number of memories $|M|$ \\
    \texttt{mem\_eff\_rank} & effective rank of $M$ \\
    \texttt{mem\_mean\_pairwise} & mean pairwise memory distance \\
    \texttt{mem\_redundancy\_at\_eps} & fraction of memories with another memory within $\varepsilon_p$ ($p{\in}\{5,10\}$) \\
    \midrule
    \multicolumn{2}{l}{\emph{Query reachability (6)}}\\
    \textbf{\texttt{q\_nn\_mem}} & \textbf{distance from $q$ to nearest memory (core reachability)} \\
    \texttt{q\_topk\_mean\_mem} & mean distance to the $\min(5,|M|)$ closest memories \\
    \texttt{q\_memory\_density}, \texttt{\_p10}, \texttt{\_p5} & \# memories within $\varepsilon_{25}$, $\varepsilon_{10}$, and $\varepsilon_5$ of $q$ \\
    \texttt{q\_nn\_turn} & distance from $q$ to nearest turn \\
    \midrule
    \multicolumn{2}{l}{\emph{Conversation structure (5, covariate control)}}\\
    \texttt{n\_turns}, \texttt{n\_sessions} & counts of turns / sessions \\
    \texttt{turn\_silhouette} & cosine silhouette of $T$ under session labels \\
    \texttt{intra\_inter\_ratio} & mean intra- / inter-session turn distance \\
    \texttt{effective\_clusters} & BIC-chosen GMM component count over $T$ \\
    \bottomrule
  \end{tabular}
\end{table}

\section{Reproducibility details}
\label{app:repro}

\paragraph{Implementations and row coverage.}
Sem-RAG is the harness store \texttt{rag-mem}, Ext-Dense is
\texttt{redis-instruct}, Ext-Hybrid is \texttt{remis-instruct-full},
Obs-Extract is \texttt{mastra-om}, and Svc-Mem is \texttt{langmem}.
The latter is a local library integration, rather than a managed service.
Rows are Sem-RAG 412, Ext-Dense 456, Ext-Hybrid 500, Obs-Extract 499,
and Svc-Mem 500. The 133 missing rows relative to $5\times500$
reflect incomplete logged answer/judge coverage.
Missingness need not be random, so run accuracies are not a fully paired
system comparison. The predictor table includes 500 distinct question IDs.
Per-type row/distinct-question counts are:
single-session-assistant 265/56, temporal-reasoning 623/133,
single-session-user 333/70, single-session-preference 137/30,
multi-session 639/133, and knowledge-update 370/78.

For each example, the feature code deduplicates the memory strings in its
answer-log records and treats their union as $M$. It does not reconstruct the
exact answer-model prompt. Claims about the complete backend store or actual
prompt inclusion would require a separate audit.

\paragraph{Classifier and preprocessing.}
We use scikit-learn \texttt{LogisticRegression}, L2 regularization,
$C=1$, \texttt{solver=lbfgs}, \texttt{max\_iter=2000}, and no class weighting.
\texttt{StratifiedGroupKFold} uses five folds, shuffling, and seed 0.
Each training fold fits its own standard scaler.
Undefined entries are mean-imputed column-wise before splitting.
There are 2{,}495 such entries: five features undefined for each of
499 single-summary Obs-Extract rows.
Columns with no finite values are filled with zero.
Per-type results fit separate models on the corresponding rows.
The difficulty baseline uses type one-hot encoding, character count,
and PCA-64 of question embeddings. PCA uses all unique question texts
without labels, including held-out covariates.
Future evaluation should fit imputation and PCA within each fold and
reserve an independent final test set.

\paragraph{Feature choices and computation.}
Query top-$k$ uses $k=5$.
Density thresholds are the 25th, 10th, and 5th percentiles of pairwise turn distances.
Coverage and redundancy use the 10th and 5th percentiles because the
25th-percentile versions saturated in exploratory analysis.
The conversation GMM selects $K\in\{1,\ldots,\min(10,|T|)\}$ by BIC,
with diagonal covariance, seed 0, and \texttt{reg\_covar}$=10^{-4}$.
Memory effective rank is exponential entropy of normalized singular
values~\citep{roy2007effective}.
Silhouette measurements use cosine distances~\citep{rousseeuw1987silhouettes}.
Figure~\ref{fig:manifold} uses t-SNE only for display~\citep{vandermaaten2008tsne}.
Embedding inputs are capped at 24{,}000 characters; local models also apply
their tokenizer's context limits. We use their default encoding configuration.
Embedding computation is additional work unless those representations are
already retained. Pairwise-distance features have quadratic time or memory
cost in the relevant set size.
We do not provide complete runtime, hardware, or API-cost accounting.

\paragraph{Uncertainty and additional checks.}
Most intervals resample question groups 1{,}000 times from fixed out-of-fold
predictions, using seed 0.
The new single-session-assistant paired comparison uses the same five folds
and 1{,}000 question-group bootstrap draws.
Some original pooled baseline differences use 2{,}000 draws.
Intervals exclude classifier refitting and exploratory feature-selection uncertainty.
The embedding comparison reports sample standard deviation across four models.
The pooled feature-block differences use 1{,}000 paired question-group draws
from the same out-of-fold predictions as Table~\ref{tab:decomp}.
Leave-one-system-out fits use L2 logistic regression, $C=1$,
and \texttt{max\_iter=3000}, training on four systems and testing the fifth.
Question identities overlap between training and test systems.

\paragraph{Generation, judging, and prompts.}
LongMemEval runs use the \texttt{gpt-4o} alias for generation and judging.
They were logged between March 31 and April 9, 2026.
The context-selection intervention uses \texttt{gpt-4o-mini} and
\texttt{gpt-4o} answer aliases, with \texttt{gpt-4o-mini} judging.
Answer and judge calls use temperature 0.
Answer prompts vary by implementation, supplying memories as context
and the question as a user message.
LongMemEval judging uses question-type-specific prompts.
Dated model snapshots were not retained, so exact replay is not guaranteed.
The linked analysis release supports offline reproduction from derived numerical
artifacts. It excludes the original multi-backend harness, raw textual logs, and
embedding caches. Memory construction, generation, and judging are not rerun.

\paragraph{LoCoMo procedure.}
We use 10 conversations and 1{,}986 questions:
single-hop 841, multi-hop 282, temporal 321, open-domain 96, adversarial 446.
Generation and judging use \texttt{gpt-4o-mini} at temperature 0.
The judge uses the harness's generic answer-correctness prompt.
Category-table labels come from the full-conversation condition
(\texttt{rich}). The fixed score is nearest-turn similarity in original space.
No classifier is trained on LoCoMo labels for that table.
Labels serve evaluation and the offline mean-distance gap.
The intervention selects 15 source turns per query.
Appropriate uncertainty would resample conversations, but category
intervals were not calculated.

\paragraph{Data sources and LLM assistance.}
We use released LongMemEval and LoCoMo data and credit their
creators~\citep{wu2025longmemeval,maharana2024locomo}.
Source repositories are
\url{https://github.com/xiaowu0162/LongMemEval} and
\url{https://github.com/snap-research/locomo}.
LoCoMo is distributed under
\href{https://creativecommons.org/licenses/by-nc/4.0/}{CC BY-NC 4.0};
we use it for noncommercial research.
A complete asset-license inventory is not supplied in this manuscript.
LLMs serve as memory extractors, answer generators, and judges.
LLM-based tools also assisted with analysis code and manuscript revision.
Reported results refer to logged experiments or the stated reanalysis
of their feature tables.

\end{document}